\documentclass[10pt,twocolumn,letterpaper]{article}

\usepackage{cvpr}
\usepackage{times}
\usepackage{epsfig}
\usepackage{graphicx}
\usepackage{amsmath}
\usepackage{amssymb}

\usepackage{bm}
\usepackage{array}
\newcolumntype{L}[1]{>{\raggedright\let\newline\\\arraybackslash\hspace{0pt}}m{#1}}
\newcolumntype{C}[1]{>{\centering\let\newline\\\arraybackslash\hspace{0pt}}m{#1}}
\newcolumntype{R}[1]{>{\raggedleft\let\newline\\\arraybackslash\hspace{0pt}}m{#1}}

\usepackage[breaklinks=true,bookmarks=false,colorlinks=true]{hyperref}

\cvprfinalcopy 

\def\cvprPaperID{956} 

\ifcvprfinal\fi
\begin{document}

\title{AMC: Attention guided Multi-modal Correlation Learning for Image Search}

\author{Kan Chen$^1$\qquad Trung Bui$^2$\qquad Chen Fang$^2$\qquad Zhaowen Wang$^2$\qquad Ram Nevatia$^1$\\
$^1$University of Southern California\qquad $^2$Adobe Research\\
{\tt\small kanchen@usc.edu, \{bui, cfang, zhawang\}@adobe.com, nevatia@usc.edu}
}

\maketitle

\begin{abstract}
   Given a user's query, traditional image search systems rank images according to its relevance to a single modality (e.g., image content or surrounding text).
   Nowadays, an increasing number of images on the Internet are available with associated meta data in rich modalities (e.g., titles, keywords, tags, etc.), which can be exploited for better similarity measure with queries. 
   In this paper, we leverage visual and textual modalities for image search by learning their correlation with input query.
   According to the intent of query, attention mechanism can be introduced to adaptively balance the importance of different modalities.
   We propose a novel Attention guided Multi-modal Correlation (AMC) learning method which consists of a jointly learned hierarchy of intra and inter-attention networks.
   Conditioned on query's intent, intra-attention networks (i.e., visual intra-attention network and language intra-attention network) attend on informative parts within each modality; a multi-modal inter-attention network promotes the importance of the most query-relevant modalities.
   In experiments, we evaluate AMC models on the search logs from two real world image search engines and show a significant boost on the ranking of user-clicked images in search results. 
   Additionally, we extend AMC models to caption ranking task on COCO dataset and achieve competitive results compared with recent state-of-the-arts.\vspace{-3.5mm}
\end{abstract}

\section{Introduction}
Image search by text is widely used in everyday life (\emph{e.g.}, search engines, security surveillance, mobile phones). 
Given a textual query, image search systems retrieve a set of related images by the rank of their relevance. Learning this relevance, \emph{i.e.},  correlation between query and image, is key to the system's utility.

\begin{figure}[h]
\includegraphics[width=2.9in]{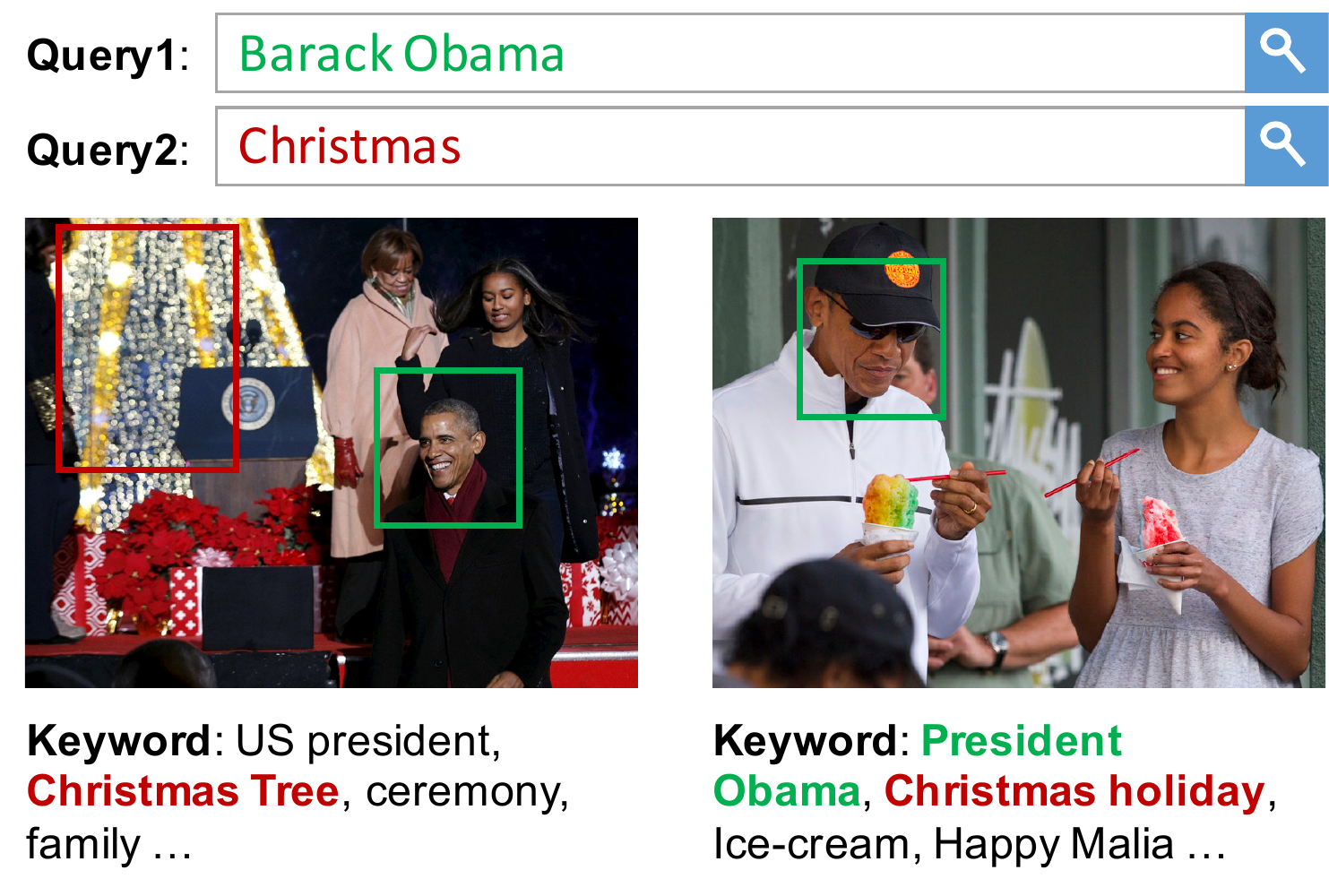}
\centering
\caption{For different queries, it is helpful to select query-dependent information within and cross rich image-related modalities available on the Internet. Bounding boxes and highlighted keywords correspond to different queries' intent by their colors.}\label{fig: intro}\vspace{-3.5mm}
\end{figure}

To measure the correlation between query and image, typically a shared latent subspace is learned for query's text modality and a single image-related modality (\emph{e.g.}, visual contents, surrounding text). 
Traditional image search engines~\cite{chowdhury2010introduction,robertson1995okapi} match queries with text or tags associated with images. 
DSSM~\cite{huang2013learning} learns an embedding subspace to measure the correlation between document-related text modality and query's text modality using deep learning. 
On the other hand, cross-modal methods \cite{yao2015learning,gordo2016deep,radenovic2016cnn,gan2016learning} learn a subspace to better measure correlation between query's text modality and image's visual modality.
In recent years, multiple image-related modalities are becoming widely available online (\emph{e.g.}, images on social networks are typically posted with captions and tags, followed by friends' comments).
Text matching and cross-modal methods are suboptimal due to their focus on only single image-related modality.
As shown in Fig~\ref{fig: intro}, image content can provide detailed visual information (\emph{e.g.}, color, texture) of objects while keywords can offer abstract concepts (\emph{e.g.}, scene description) or external background information (\emph{e.g.}, people's identities). 
Different modalities describe images from different views, which together provide information in a more comprehensive way.
It benefits to learn a subspace to measure the correlation between query's text modality and image-related modalities, \emph{i.e.}, multi-modal correlation.

There is a major challenge in learning this subspace: not all modalities are equally informative due to the variation in query's intent.
To overcome this problem, we introduce an attention mechanism to adaptively evaluate the relevance between a modality and query's intent.
For the image search task, we consider two kinds of attention mechanisms.
First, there is query-unrelated information within each modality (\emph{e.g.}, background regions in images, keyword ``Ice-cream'' for query2 ``Christmas'' in Fig~\ref{fig: intro}); 
an image search system should attend on the most informative parts for each modality (\emph{i.e.}, intra-attention). 
Second, different modalities' contributions vary for different queries; 
an image search system should carefully balance the importance of each modality according to query's intent (\emph{i.e.}, inter-attention).

To address the aforementioned issues, we propose a novel Attention guided Multi-modal Correlation (AMC) learning method. AMC framework contains three parts: visual intra-attention network (VAN), language intra-attention network (LAN) and multi-modal inter-attention network (MTN). 
VAN focuses on informative image regions according to query's intent by generating a query-guided attention map.
LAN learns to attend on related words by learning a bilinear similarity between each word in language modality and query.
MTN is built to attend between different modalities. 
Finally, the correlation between query and image-related modalities is calculated as the distance between query embedding vector and a multi-modal embedding vector in the learned AMC space. 

To validate the AMC framework, we choose image-related keywords as the language modality and image contents as the visual modality.
AMC models are evaluated on two datasets: Clickture dataset~\cite{yao2015learning} and Adobe Stock dataset (ASD). 
ASD is collected from Adobe Stock search engine, including queries, images, manually curated keywords and user clickthrough data.
For Clickture, we curated keywords for all images by an auto-tagging program developed internally at Adobe.
Experiments show that AMC achieves significant improvement on both datasets.
More importantly, this finding indicates that AMC can benefit from not only human curated data, but also information generated by machines, which could be noisy and biased.
Moreover, since AMC can scale to any number of modalities, it has the ability to integrate and benefit from the output of any intelligent visual analysis system.
We further evaluate AMC for caption ranking task on COCO image caption data~\cite{karpathy2015deep} with keyword set curated in the same way for Clickture~\cite{yao2015learning}.
AMC models achieve very competitive performance, even surpass the state-of-the-art method in Recall@10 metric. 

Our contributions are as follows: we propose a novel AMC learning framework to select query-dependent information within and cross different modalities. 
AMC model achieves significant improvement in image search task.
We plan to release the auto-tagged Clickture and COCO dataset upon publication. \vspace{-2.0mm}

\section{Related Work}\label{sec: related}
\textbf{Multi-modal Correlation learning.} Canonical correlation analysis (CCA)~\cite{hardoon2004canonical} learns a cross-modal embedding space to maximize the correlation between different modalities. Kernel CCA (KCCA)~\cite{fukumizu2007statistical} extends CCA by adopting a non-linear mapping for different modalities. Alternatively, Nakayama \emph{et al.} propose kernel principle component analysis with CCA (KPCA-CCA)~\cite{nakayama2010evaluation}, which generates input for CCA via non-linear KPCA method. Gong \emph{et al.}~\cite{gong2014multi} further include a third view into the CCA space by the semantics between image and tags. Similarly, partial least squares (PLS)~\cite{rosipal2006overview} aims to measure the correlation by projecting multiple sets of data into a latent space. Ngiam \emph{et al.}~\cite{ngiam2011multimodal} introduce deep multimodal learning using neural networks. Recently, Datta \emph{et al.}~\cite{datta2017multimodal} first expand to learn the correlation between query and multiple image-related modalities using a graph-based keyphrase extraction model. Compared to~\cite{datta2017multimodal}, AMC effectively learns a multimodal subspace to measure correlation between query and image-related modalities via three attention networks driven by click-through data. 

\textbf{Attention network.} Attention mechanisms have been successfully applied in many computer vision tasks, including object detection \cite{mnih2014recurrent} and fine-grained image classification \cite{lin2015bilinear}.
Jin \emph{et al.}~\cite{jin2015aligning} develop an attention-based model for image captioning task that employs an RNN to attend on informative regions in images.
Yang \emph{et al.}~\cite{yang2015stacked} and Chen \emph{et al.}~\cite{chen2015abc} apply attention networks that focus on useful regions in visual question answering (VQA) task.
Xiong \emph{et al.}~\cite{xiong2016dynamic} propose a dynamic memory network to attend on informative visual or textual modality for question answering. 
Recently, Lu \emph{et al.}~\cite{lu2016hierarchical} propose a co-attention network to focus on both visual and question modalities in VQA task. 
Compared to these methods, AMC method not only applies intra-attention networks within each modality, but also employs MTN to balance the importances of modalities based on query's intent for image search task.

\begin{figure*}[h]
\includegraphics[width=7in]{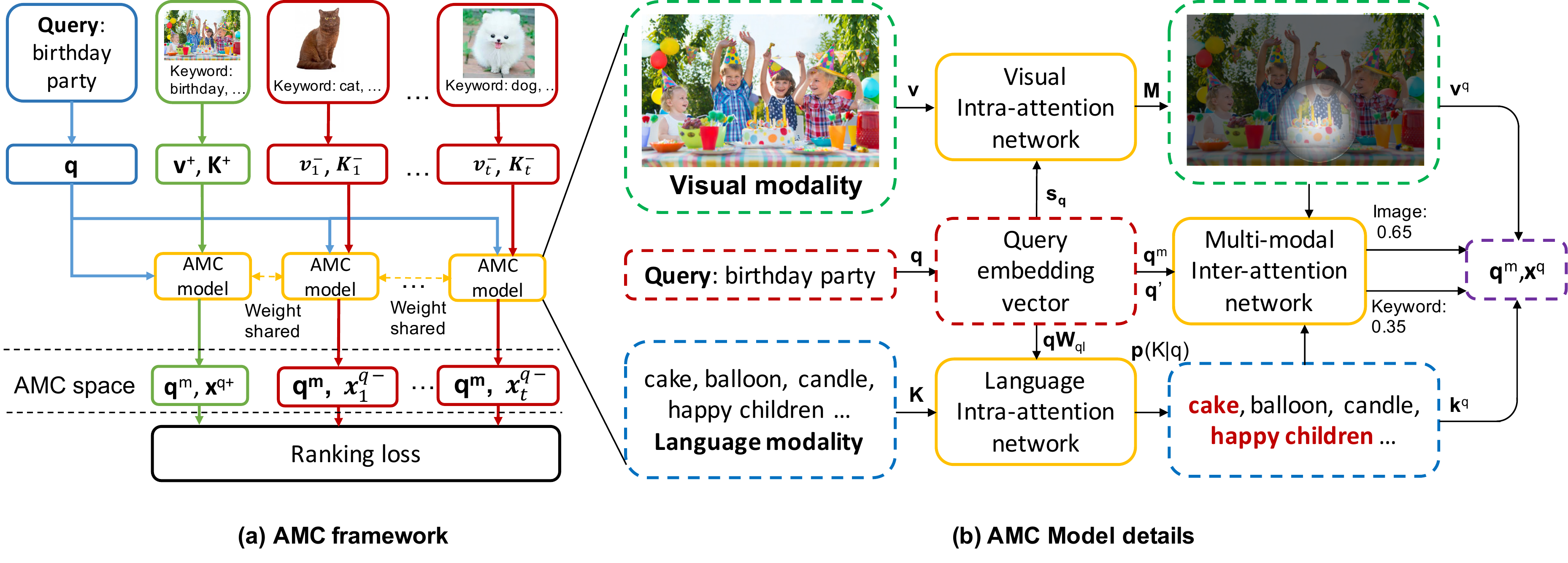}
\centering
\caption{Attention guided Multi-modal Correlation (AMC) learning framework. Left: Given a query, images and related keywords are projected to a raw embedding space. AMC model then generates a query-guided multi-modal representation for each image. The correlation between query and image is measured by the cosine distance in the AMC space. Right: AMC model consists of a visual intra-attention network (VAN), a language intra-attention network (LAN) and a multi-modal inter-attention network (MTN). VAN and LAN attend on informative parts within each modality and MTN balances the importance of different modalities according to the query's intent.}\label{fig: pipeline}
\end{figure*}

\textbf{Image and textual search.} For image search task, CCA~\cite{hardoon2004canonical} is employed to learn a subspace to maximize correlation between query and image. Ranking CCA (RCCA)~\cite{yao2015learning} refines the CCA space by learning a bilinear ranking function from click-through data. Wang \emph{et al.}~\cite{wang2014learning} apply a deep ranking model for fine-grained image search and Tan \emph{et al.}~\cite{zhao2015deep} introduce a deep ranking based hashing model. Recently, Gordor \emph{et al.}~\cite{gordo2016deep} apply a region proposal network and Radenovi{\'c} \emph{et al.}~\cite{radenovic2016cnn} adopt deep CNN features. Lynch \emph{et al.}~\cite{lynch2015images} transfer deep semantic features learned from click-through data and apply them on image search task. Compared to the approaches above, AMC method applies VAN to adaptively select informative regions within image modality based on query's intent. On the other side, for textual search task, Joachims~\cite{joachims2002optimizing} introduces click-through data for optimizing search engines. DSSM~\cite{huang2013learning} applies a deep framework to further leverage click-through data. Compared to DSSM~\cite{huang2013learning}, AMC method employs LAN to attend on query-related words.


\section{AMC Learning From Click-through Data}\label{sec: amc method}
The goal of Attention guided Multi-modal Correlation learning (AMC) method is to construct an AMC space where the correlation between query $q$ and image $x$ can be measured by the distance between query's embedding vector $\mathbf{q}^m$ and image's query-guided multi-modal representation $\mathbf{x}^q$ (superscript ``$m$'' denotes the multi-modal subspace in $\mathbf{q}^m$).
To learn the AMC space, we propose a hierarchy of intra and inter attention networks, \emph{i.e.}, visual intra-attention network (VAN), language intra-attention network (LAN) and multi-modal inter-attention network (MTN). 
In this paper, we select image-related keywords as the language modality and image visual contents as the visual modality, while the AMC space can be further extended to incorporate more image-related modalities.

We first present the AMC learning framework followed by the details of inter-attention network (MTN). Intra-attention networks (VAN and LAN) are then introduced. Finally, we illustrate how to apply the learned AMC space on image search and caption ranking tasks.

\subsection{AMC learning framework}
In AMC space, the correlation between a query $q$ and an image $x$ is measured by the cosine distance $\langle\mathbf{q}^m,\mathbf{x}^q\rangle$, where $\mathbf{q}^m\in\mathbb{R}^d$ is the embedding vector of $q$ and $\mathbf{x}^q\in\mathbb{R}^d$ is the multi-modal representation of $x$ conditioned on query's intent. 
To learn the AMC space, we sample $\mathcal{N}$ tuples in the form $[q, (x^+, K^+),(x^-_1, K^-_1),$ $(x^-_2,K^-_2),... ,(x^-_t,K^-_t)]$ from click-through data. 
Each tuple consists of a query $q$, a positive image $x^+$ with its keyword set $K^+$ and $t$ negative images $x^-_i$ with their keyword sets $K^-_i$. 
Given the query $q$ in a tuple, the positive image $x^+$ has the highest number of clicks. Similar to~\cite{yao2015learning}, we adopt a common ranking loss function as the objective:
\begin{equation}\label{equ: amc loss}
\begin{split}
&\arg\min_{\theta} \sum_{i=1}^\mathcal{N}\mathcal{L}_{\theta}(q_i, \{x^+_i, K^+_i\}, \{x^-_{ij},K^-_{ij}\}^t_{j=1})\\
&\mathcal{L}_{\theta} = \sum_{j=1}^{t}\max(0, \alpha-\langle \mathbf{q}^m_i, \mathbf{x}^{q+}_i\rangle+\langle \mathbf{q}^m_i, \mathbf{x}^{q-}_{ij}\rangle)
\end{split}
\end{equation}
where $\theta$ denotes the model's parameters to be optimized and $\alpha$ is the margin between positive and negative samples.

To learn the query's embedding $\mathbf{q}^m$ and query-guided multi-modal representation $\mathbf{x}^q$ for image $x$, we propose a multi-modal inter-attention network (MTN) to attend on informative modalities. 
The inputs of MTN are query-guided single modality embeddings produced by intra-attention networks. 
Specifically, intra-attention networks consist of a visual intra-attention network (VAN) and a language intra-attention network (LAN).
For visual modality, VAN focuses on useful regions in image contents and generates a query-guided visual embedding $\mathbf{v}^q\in\mathbb{R}^d$; for language modality, LAN filters out unrelated words and generates a query-guided language embedding $\mathbf{k}^q\in\mathbb{R}^d$. 
The AMC framework is trained in an end-to-end way by integrating VAN, LAN and MTN  (Fig~\ref{fig: pipeline}).

For simplicity, we denote the input feature for query $q$ as $\mathbf{q}\in\mathbb{R}^{d_q}$. Each image $x$ is represented as a $r\times r$ feature map $\mathbf{v}\in\mathbb{R}^{r\times r\times d_v}$. The input feature matrix for keyword set $K$ is denoted as $\mathbf{K} = \{\mathbf{k}_{1},\mathbf{k}_{2},...,\mathbf{k}_{n}\}^\top\in\mathbb{R}^{n\times d_k}$, where $n$ is the keyword set size and $\mathbf{k}_{j}$ is the $j$-th keyword's feature vector of image $x$. $d_q$, $d_k$ and $d_v$ are the feature dimensions for query, keyword and image respectively.

\subsection{Multi-modal inter-attention network (MTN)}\label{sec: mtn}
MTN generates the embedding $\mathbf{q}^m$ of query by projecting query's input feature $\mathbf{q}$ into AMC space through a non-linear transform. 
\begin{equation}\label{equ: q_m}
\mathbf{q}^m=f(\mathbf{W}_{qm}\mathbf{q}+\mathbf{b}_{qm})
\end{equation}
where $\mathbf{W}_{qm}\in\mathbb{R}^{d_q\times d}, \mathbf{b}_{qm}\in\mathbb{R}^d$ are the linear transformation matrix and bias vector to be optimized. $f(.)$ is a non-linear activation function.
Besides, MTN encodes query's intent $\mathbf{q}^\prime$ using another similar transform in Eq~\ref{equ: q_m}. 
Conditioned on the query's intent, the correlation of embeddings $[\mathbf{v}^q, \mathbf{k}^q]$ produced by VAN and LAN is calculated as: 
\begin{equation}\label{equ: mtn cos}
[c_v, c_k] = \langle\mathbf{q}^\prime, [\mathbf{v}^q, \mathbf{k}^q]\rangle,\text{\ \ } \mathbf{q}^\prime=f(\mathbf{W}_{qm}^\prime\mathbf{q}+\mathbf{b}_{qm}^\prime)
\end{equation}  
$[c_v, c_k]$ denotes the correlation of visual and language modality. $\langle.,.\rangle$ is the cosine distance measurement.  
$f(.)$ is a non-linear activation function. $\mathbf{W}_{qm}^\prime, \mathbf{b}_{qm}^\prime$ are variables to be optimized. 
MTN then re-weights the visual and language modalities based on their probabilities conditioned on the input query's intent (\emph{e.g.}, in Fig~\ref{fig: pipeline}, the relevance scores for visual modality (``Image'') and language modality (``Keyword'') are 0.65 and 0.35, indicating visual modality is more relevant than language modality for query ``birthday party''). 
The conditional probability for each modality is measured based on the correlation in Eq~\ref{equ: mtn cos}. 
The final multi-modal embedding $\mathbf{x}^q\in\mathbb{R}^d$ in the AMC space is:
\begin{equation}\label{equ: multi-modal embed}
\mathbf{x}^q = p_v\mathbf{v}^q+p_k\mathbf{k}^q,\text{\ \ } [p_v, p_k] = \sigma([c_v, c_k])
\end{equation}
where $\sigma(.)$ is a softmax function. $\mathbf{x}^q$ encodes the useful information from different modalities conditioned on the input query's intent.

\subsection{Visual intra-attention network (VAN)}\label{sec: van}
VAN takes query $q$'s input feature $\mathbf{q}$ and image $x$'s feature map $\mathbf{v}$ as input. 
It first projects image feature map $\mathbf{v}$ into a $d$-dimension raw visual subspace by a 1x1 convolution kernel $\mathbf{W}_v\in\mathbb{R}^{d_v\times d}$. The projected image feature map is denoted as $\mathbf{v}^\prime\in\mathbb{R}^{r\times r\times d}$.
Similar to~\cite{chen2015abc}, VAN generates a query-guided kernel $\mathbf{s}_q$ from query embedding vector $\mathbf{q}$ through a non-linear transformation. 
By convolving the image feature map with $\mathbf{s}_q$, VAN produces a query-guided attention map $\mathbf{M}$:
\begin{equation}\label{equ: vis att map}
\mathbf{M} = \sigma(\mathbf{s}_q * \mathbf{v}^\prime),\text{\ \ } \mathbf{s}_q = f(\mathbf{W}_{qs}\mathbf{q}+\mathbf{b}_{qs})
\end{equation}
where $f(.)$ is a non-linear activation function. $\sigma(.)$ is a softmax function and ``*'' is the convolution operator. $\mathbf{W}_{qs}, \mathbf{b}_{qs}$ are the linear transformation matrix and bias vector that project query embedding vector $\mathbf{q}$ from the language space into the kernel space. The generated attention map is of the same resolution as image feature map $\mathbf{v}^\prime$ ($r\times r$). Each element in the attention map represents the probability of the corresponding region in image $x$ being informative conditioned on the intent of query $q$.

VAN then refines the raw visual subspace through re-weighting each location of projected image feature map $\mathbf{v}^\prime$ by the corresponding conditional probability in the attention map $\mathbf{M}$ via element-wise production. The query-guided visual embedding vector $\mathbf{v}^{q}\in\mathbb{R}^{d}$ for image $x$ is generated by average pooling of the re-weighted image feature map:
\begin{equation}\label{equ: vis embed}
\mathbf{v}^q = \text{AvgPool}(\mathbf{M}\odot\mathbf{v}^\prime)
\end{equation}
where ``AvgPool'' is the average pooling operation and $\odot$ represents element-wise production. 

\subsection{Language intra-attention network (LAN)}\label{sec: lan}
LAN takes query input feature vector $\mathbf{q}$ and keyword set feature matrix $\mathbf{K}$ as inputs. It first projects query $\mathbf{q}$ and keywords $\mathbf{K}$ into a raw language subspace by linear projections. Similar to~\cite{yao2015learning}, the correlation between input query and keywords is measured in a bilinear form:
\begin{equation}\label{equ: keyword similarity}
\bm{s}(q, K, \mathbf{W}_{ql}, \mathbf{W}_{kl}, \mathbf{W}_{l}) = (\mathbf{q}\mathbf{W}_{ql})\mathbf{W}_l(\mathbf{K}\mathbf{W}_{kl})^\top
\end{equation} 
where $\mathbf{W}_{ql}\in\mathbb{R}^{d_q\times d}$ and $\mathbf{W}_{kl}\in\mathbb{R}^{d_k\times d}$ are transformation matrices that project query $\mathbf{q}$ and keywords $\mathbf{K}$ into the raw subspace. $\mathbf{W}_l\in\mathbb{R}^{d\times d}$ is the bilinear similarity matrix. 
Since $d<d_q, d<d_k$, $\{\mathbf{W}_{ql}, \mathbf{W}_{kl}, \mathbf{W}_l\}$ are like an SVD decomposition of the overall $d_q\times d_k$ bilinear matrix.
LAN then refines the raw language subspace by re-weighting each keyword embedding vector by their probability conditioned on the query's intent. This probability is measured based on the similarity between query $\mathbf{q}$ and keywords $\mathbf{K}$ in Eq~\ref{equ: keyword similarity}. The refined language embedding $\mathbf{k}^q\in\mathbb{R}^d$ for keyword set $K$ is calculated as
\begin{equation}
\mathbf{k}^q = \bm{p}(K | q)^\top\mathbf{K}\mathbf{W}_{kl},\text{\ \ } \bm{p}(K|q) = \sigma(\bm{s}(q, K))
\end{equation}
where $\bm{s}(q, K)$ is the correlation between query and keywords calculated in Eq~\ref{equ: keyword similarity}. $\sigma(.)$ is the softmax function. $\bm{p}(K|q)$ is the probability of each keyword being informative conditioned on the query's intent. 

\subsection{Applications of AMC space} 
The learned AMC space can be applied directly on two tasks: image search and caption ranking.
For image search, we first calculate the input query $q$'s embedding vector $\mathbf{q}^m$ in the learned AMC space. 
We then generate the multi-modal representations $\{\mathbf{x}^q\}$ for all the images in the dataset. 
The images are ranked based on their relevance to the input query, which is measured by the cosine distance between $\mathbf{q}^m$ and $\{\mathbf{x}^q\}$. 

For caption ranking, we adopt another objective function in~\cite{kiros2015skip} during training for fair comparison:
\begin{equation}\label{equ: cap rank loss}
\begin{split}
\mathcal{L}_{\theta} = &\sum_{\mathbf{x}}\sum_k\max\{0, \alpha-\langle\mathbf{x}^q, \mathbf{q}^m\rangle+\langle\mathbf{x}^q, \mathbf{q}^m_k\rangle\} \\
+&\sum_{\mathbf{q}}\sum_k\max\{0, \alpha-\langle\mathbf{x}^q, \mathbf{q}^m\rangle+\langle\mathbf{x}^q_k, \mathbf{q}^m\rangle\}
\end{split}
\end{equation}
where $\mathbf{q}^m$ is the caption embedding vector and $\mathbf{x}^q$ is the multi-modal embedding vector of image $x$. The subscript $k$ indicates negative embeddings for current caption-image (keyword) pairs and $\langle.,.\rangle$ is the cosine distance measurement.
Given a query image $x$ and related modalities, we first calculate all candidate captions' embedding vectors $\mathbf{q}^m$ in the learned AMC space. 
The multi-modal representations for images conditioned on the caption's intent $\{\mathbf{x}^q\}$ are then generated by the AMC model.
Finally, each caption $q$ is ranked based on the correlation between $\mathbf{q}^m$ and $\mathbf{x}^q$.

We choose the rectified linear unit (ReLU) as the activation function $f(.)$. 
AMC model's parameters $\theta$ consist of the variables: $\{\mathbf{W}_v, \mathbf{W}_{qs}, \mathbf{b}_{qs}, \mathbf{W}_{ql}, \mathbf{W}_{qm}^\prime, \mathbf{b}_{qm}^\prime,$ $\mathbf{W}_{kl}, \mathbf{W}_l, \mathbf{W}_{qm}, \mathbf{b}_{qm}\}$. 
We apply adam~\cite{kingma2014adam} algorithm to train the AMC framework in an end-to-end way.


\section{Dataset}\label{click-keyword-dataset}
\begin{figure}[h]
\includegraphics[width=3.3in]{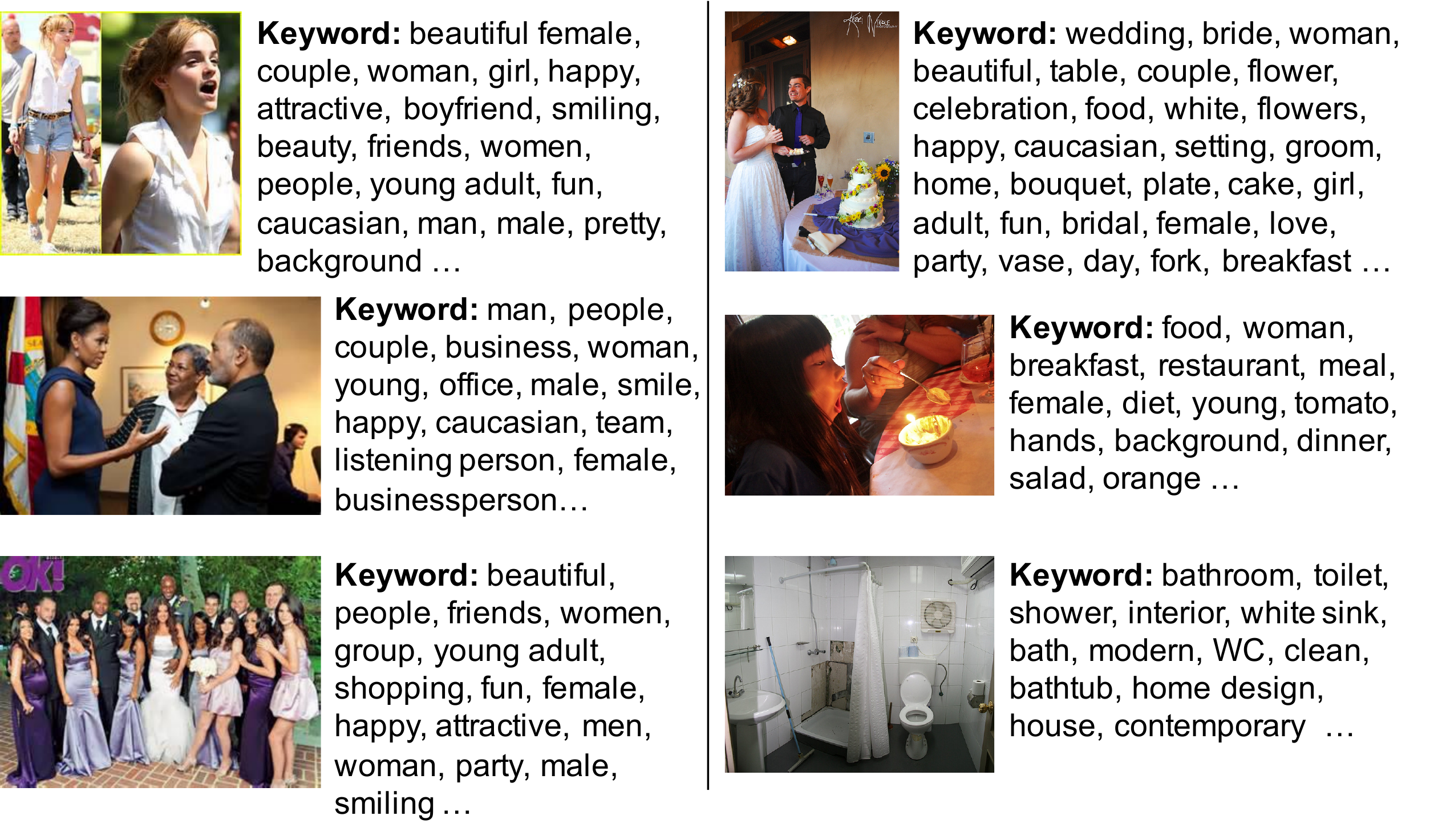}
\centering
\caption{Images with keywords in Clickture~\cite{yao2015learning} (left) and COCO image caption dataset~\cite{lin2014microsoft} (right). Since each image is associated with $\sim$100 keywords, not all keywords are listed.}\label{fig: key_data}\vspace{-3.5mm}
\end{figure}

\textbf{Keyword datasets\footnote{Available in \url{https://github.com/kanchen-usc/amc_att}}.} We curated two keyword datasets for Clickture~\cite{yao2015learning} and COCO~\cite{lin2014microsoft} by an auto-tagging system. 
Basically, given a query image, the system first searches similar images from a commercial image database using a k-NN ranking algorithm.
Then the query image's keywords are generated based on a tag voting program among the keywords associated with the images from k-NN ranking results. 
The Clickture keyword dataset has over 50k unique keywords. The average size of keyword sets is 102 (minimum is 71 and maximum is 141).
There are over 26k unique keywords in COCO keyword dataset. The average size of keyword sets is 102 (minimum size is 99 and maximum size is 104). 
Compared to COCO object labels in COCO dataset~\cite{lin2014microsoft} which have only 91 object categories, our keyword dataset is much richer and more diverse. Besides, the keyword dataset contains multi-word phrases, upper and lower cases, which simulates the noisy keywords collected from real-world websites (Fig~\ref{fig: key_data}).

\textbf{Adobe Stock Dataset (ASD).} We collect clickthrough data from the log files in Adobe Stock\footnote{\url{https://stock.adobe.com}}. ASD contains 1,555,821 images and 1,941,938 queries, which form 3,485,487 $\{query, image, click\}$ triads. In addition, each image is associated with a set of keywords with an average size of 53. There are over 27k unique keywords in ASD. We evaluate AMC models for image search task on ASD.

\textbf{Clickture dataset~\cite{yao2015learning}} is composed of two parts: the training and development (dev) sets. The training set contains 23.1M \{$query, image, click$\} triplets. The dev set consists of 79,926 $\langle query, image\rangle$ pairs generated from 1000 queries. We evaluate AMC models for image search task on Clickture with our keyword dataset.

\textbf{COCO Image Caption dataset~\cite{karpathy2015deep} (CIC)}. COCO image dataset~\cite{lin2014microsoft} has 82,783 images for training and 413,915 images for validation. CIC shares the same training set with COCO. The validation set of CIC is composed of 1,000 images sampled from the COCO validation images, and the test set of CIC consists of 5,000 images sampled from the COCO validation images which are not in the CIC validation set. Each image in CIC is associated with 5 candidate captions. Same as \cite{karpathy2015deep}, we evaluate AMC model on the first 1,000 images for caption ranking on the CIC test set with our curated keywords.

\section{Experiments}\label{experments}
We evaluate our approach on Clickture~\cite{hua2013clickage} and Adobe Stock Dataset (ASD) for image search task, and COCO Image Caption dataset~\cite{lin2014microsoft} (CIC) for caption ranking task.

\subsection{Multi-modal image retrieval}\label{sec: multi-modal ir}
\textbf{Experiment setup.} For the visual modality, we divide an input image into 3x3 grids, and apply a pre-trained 200-layer ResNet~\cite{He2015} to extract image feature for each grid. Thus, each image is represented as a 3x3x2048 feature map ($r=3, d_v=2048$). For models without VAN, we extract global image features, and represent each image as a 2048 dimension (2048D) feature vector.  For the query and keyword modalities, we remove stop words and uncommon words in the raw data, convert all words to lowercase, and tokenize each word to its index in the corresponding dictionary. The dictionary sizes for keyword modality in Clickture and ASD are 50234 and 27822. The dictionary sizes for the query modality in Clickture and ASD are 85636 and 17388. We randomly split ASD into three parts: 70\% for training, 10\% for validation and 20\% for testing.

\textbf{Compared approaches.} We compare the following approaches for performance evaluation:

(1) Ranking Canonical Correlation Analysis~\cite{yao2015learning} (RC- 
CA) ranks images based on a bilinear similarity function learned from clickthrough data. 
We adopt Resnet~\cite{He2015} features for RCCA framework which achieves better performance compared to ~\cite{yao2015learning} using AlexNet~\cite{krizhevsky2012imagenet} features.

(2) Multimodal Bilinear Pooling (MB) combines visual and language modalities by an outer production layer. Compared to multimodal compact bilinear pooling (MCB) model~\cite{park2016multimodal}, we drop the sketch count projection to avoid loss of information from original modalities.

(3) Deep structured semantic model~\cite{huang2013learning} (DSSM) learns a subspace to measure the similarity between text modality and queries for document retrieval using a deep learning framework. 
We build similar structures which takes single image-related modality for image search task. 
Specifically, image modality (DSSM-Img) and keyword modality (DSSM-Key) are evaluated. 

\textbf{Attention networks and AMC models.} We compare different attention networks as follows:

(1) VAN attends on informative regions in the image modality based on the query's intent.

(2) LAN selects useful words in the keyword modality based on the query's intent.

(3) Late fusion network (LF) first calculates the similarity scores between the input query and each modality. To represent the final correlation between the query and image-related modalities, LF then combines these similaritiy scores by a linear transformation.

(4) MTN balances the importance of different modalities based on the query's intent.
\setlength{\tabcolsep}{2.5pt}
\begin{table}[t]
  \centering
  \begin{tabular}{|l|C{0.7cm}|C{0.7cm}|c|c|C{0.7cm}|c|} \hline
    Approach & Img& Key&VAN&LAN& LF &MTN\\ \hline
    MB~\cite{park2016multimodal} & \checkmark & \checkmark & & & & \\ \hline
    DSSM-Key~\cite{huang2013learning} & & \checkmark & & & & \\ \hline    
    DSSM-Img~\cite{huang2013learning} & \checkmark & & & & & \\ \hline
    RCCA~\cite{yao2015learning} & \checkmark & & & & & \\ \hline        
    Img\textsubscript{ATT} & \checkmark & & \checkmark & & & \\ \hline
    Key\textsubscript{ATT} & & \checkmark & & \checkmark & &\\ \hline
    Img\textsubscript{ATT}-Key\textsubscript{ATT}-LF & \checkmark & \checkmark & \checkmark & \checkmark & \checkmark &  \\ \hline
    AMC Full & \checkmark & \checkmark & \checkmark & \checkmark & & \checkmark \\ \hline
  \end{tabular}
  \vspace{2.0mm}
\caption{Different models evaluated on Clickture and ASD. Language intra-attention network (LAN) is applied on keyword modality. Visual intra-attention network (VAN) is applied on image modality.  Late fusion (LF) and multi-modal inter-attention networks (MTN) are applied on multi-modalities.}\label{tab:amc_model_img_retrieve}
\end{table}

Different models evaluated on Clickture dataset and ASD are listed in Table~\ref{tab:amc_model_img_retrieve}, with details on adopted modalities and attention networks.

\textbf{Training details.} On Clickture dataset, we sample one negative tuple $(v^-, K^-)$ ($t=1$) while on ASD, we sample 3 negative tuples ($t=3$). Same as~\cite{yao2015learning}, the dimension of embedding vectors in all modalities is 80 ($d=80$). The batch size is set to 128. We set margin $\alpha=1$ in Eq~\ref{equ: amc loss}.

\textbf{Evaluation metrics.}
For Clickture dataset, we calculate NDCG@k score~\cite{yao2015learning} for top $k\in\{5, 10, 15, 20, 25\}$ ranking results for an input query. 
We exclude queries with ranking list's size less than $k$ for calculating NDCG@k score. 
The final metric is the average of all queries' NDCG@k in the Clickture dev set. 
We further compare different models' performance under P@5 (precision at top 5 results), P@k, MAP and MRR metrics, whose details are described in~\cite{mcfee2010metric}. 
ROC curves and Area Under Curve (AUC) are also compared between different models on Clickture Dataset.






For ASD, we use Recall at $k$ samples (R@k) as metric. Given a rank list, R@k is the recall of positive samples (ratio of clicked images among all clicked images of the input query) among the top $k$ results. The final metric is the average of all queries' R@k in the ASD test set.

\setlength{\tabcolsep}{3.5pt}
\begin{table}[t]
  \centering
  \small
  \begin{tabular}{|l|c|c|c|c|c|} \hline
    Approach & 5 & 10 & 15 & 20 & 25 \\ \hline
    MB & 0.5643 & 0.5755 & 0.5873 & 0.5918 & 0.5991 \\ 
    DSSM-Key & 0.5715 & 0.5745 & 0.5797 & 0.5807 & 0.5823 \\       
    DSSM-Img & 0.6005 & 0.6081 & 0.6189 & 0.6192 & 0.6239 \\  
    RCCA & 0.6076 & 0.6190 & 0.6293 & 0.6300 & 0.6324 \\ \hline
    Key\textsubscript{ATT} & 0.5960 & 0.6054 & 0.6168 & 0.6204 & 0.6241 \\ 
    Img\textsubscript{ATT} & 0.6168 & 0.6233 & 0.6308 & 0.6350 & 0.6401 \\
    Img\textsubscript{ATT}-Key\textsubscript{ATT}-LF & 0.6232 & 0.6254 & 0.6344 & 0.6376 & 0.6444 \\ 
    AMC Full & \textbf{0.6325} & \textbf{0.6353} & \textbf{0.6431} & \textbf{0.6427} & \textbf{0.6467} \\ \hline
  \end{tabular}
  \vspace{1.0mm}
\caption{Performance of different models on Clickture dataset. The evaluation metrics are NDCG@5, 10, 15, 20, 25 (correspond to 2\textsuperscript{nd} to 6\textsuperscript{th} column). For $k\in\{5, 10, 10, 20, 25\}$, we exclude queries with ranking list size less than $k$ when we calculate NDCG@$k$.}\label{tab:clickture_ndcg}
\end{table}

\setlength{\tabcolsep}{3.5pt}
\begin{table}[t]
  \centering
  \begin{tabular}{|l|c|c|c|c|c|} \hline
    Approach & P@5 & P@k & MAP & MRR & AUC \\ \hline
   MB & 0.5615 & 0.6372 & 0.7185 & 0.7564 & 0.6275 \\    
    DSSM-Key & 0.5431 & 0.6756 & 0.6969 & 0.7884 & 0.5508 \\
    DSSM-Img & 0.5835 & 0.6705 & 0.7308 & 0.7773 & 0.6455 \\
    RCCA & 0.5856 & 0.6778 & 0.7332 & 0.7894 & 0.6384 \\ \hline
    AMC Full & \textbf{0.6050} & \textbf{0.7069} & \textbf{0.7407} & \textbf{0.8067} & \textbf{0.6727} \\ \hline
  \end{tabular}
  \vspace{1.0mm}
\caption{Models' performance under different metrics}\label{tab:clickture_other_metric}
\end{table}
\begin{figure}[h]
\includegraphics[width=3.0in]{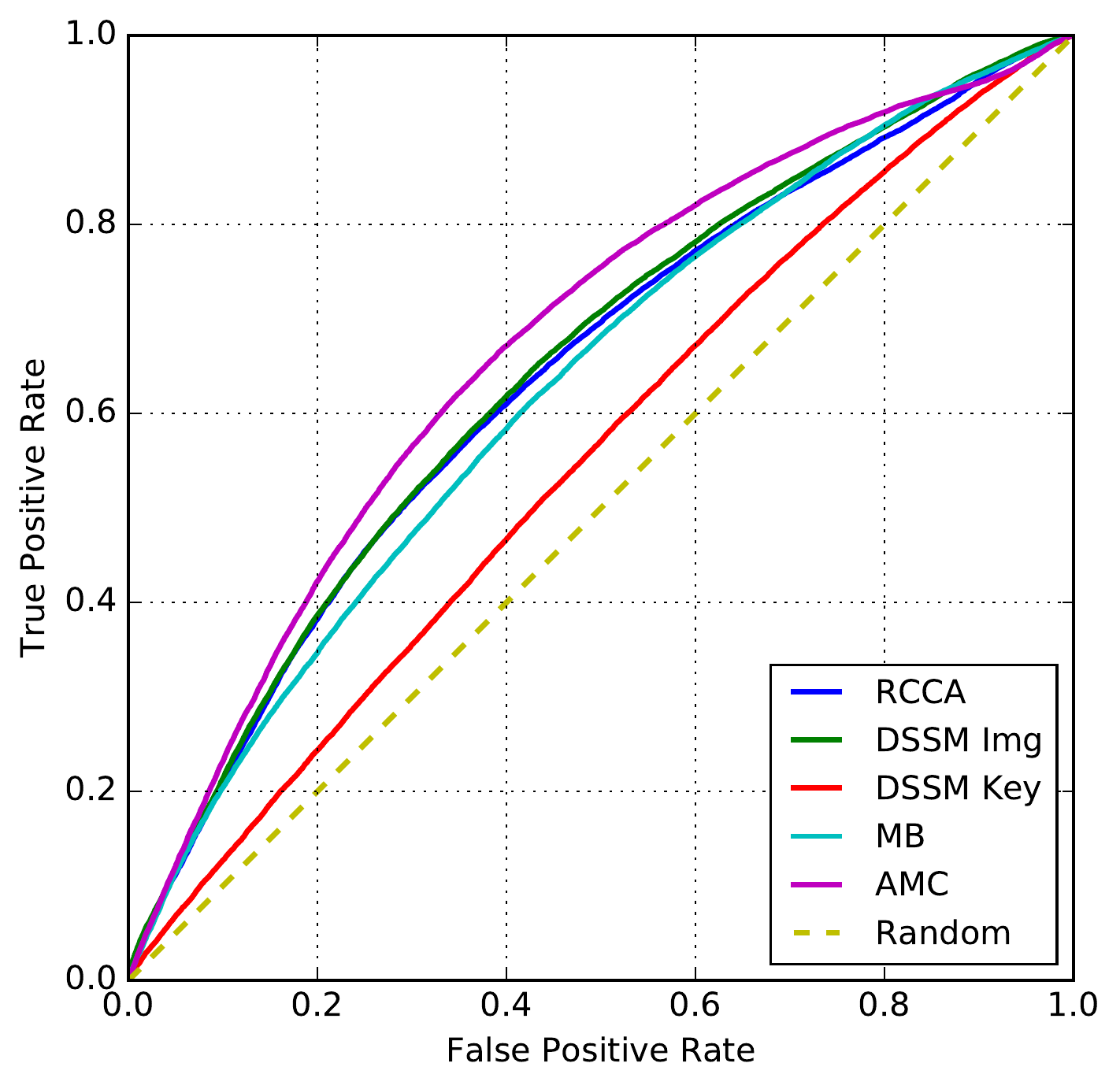}
\centering
\caption{ROC curve for different models.}\label{fig: roc}
\end{figure}

\textbf{Performance on Clickture.} The performances of different models on Clickture dataset are shown in Tables~\ref{tab:clickture_ndcg},~\ref{tab:clickture_other_metric} and Fig~\ref{fig: roc}. 
We first apply intra-attention networks on single modality models, which filters out unrelated information within each modality according to the query's intent. 
The resulting models, Key\textsubscript{ATT} and Img\textsubscript{ATT}, achieve 2.2\% and 2.6\% increase in NDCG@5 compared to DSSM-Key and DSSM-Img, respectively. Attention-guided single modality model Img\textsubscript{ATT} even beats the MB model with two modalities information in NDCG metric. 
We further applies the late fusion network (LF) on two attention-guided modalities. 
The resulting model Img\textsubscript{ATT}-Key\textsubscript{ATT}-LF achieves an additional 1\% increase in NDCG@5 compared to Img\textsubscript{ATT} and Key\textsubscript{ATT}, which validates the effectiveness of learning a multi-modal subspace to further boost the image search task. 
Finally, we apply MTN to select informative modalities based on the query's intent. 
The AMC full model achieves the state-of-the-art performance on NDCG metric, with more than 3\% increase from single modality models, and 2.5\% increase in NDCG@5 compared to RCCA model~\cite{yao2015learning}, which is $\sim$3 times of RCCA's increase compared to the previous state-of-the-art method.

\begin{figure}[h]
\includegraphics[width=3.2in]{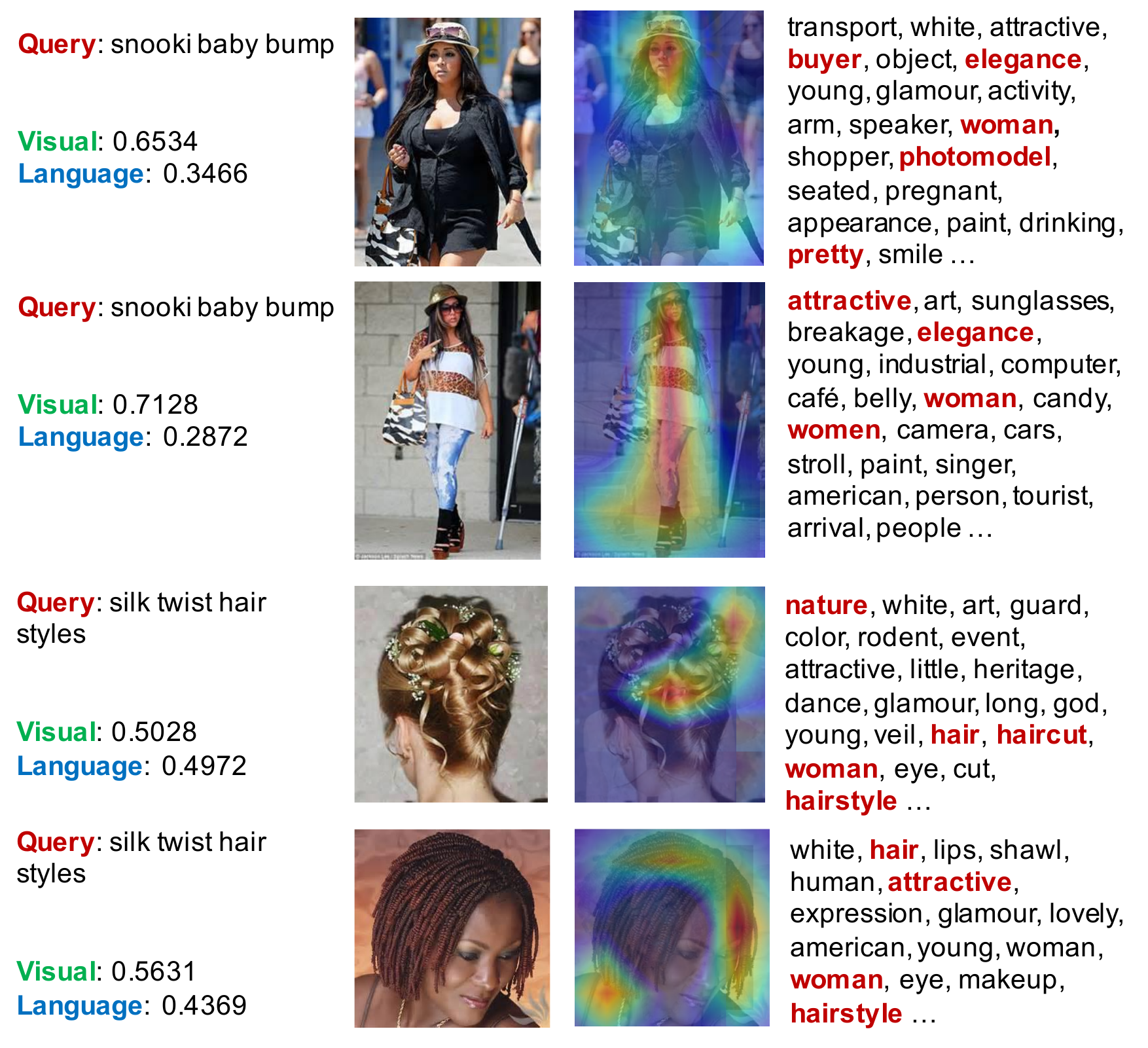}
\centering
\caption{Visualization of AMC model's VAN, LAN and MTN results. First column: Input query and importance of visual and language modalities produced by MTN. Second and third columns: original images and query-guided attention maps produced by VAN. Fourth column: Some keywords highlighted by LAN.}\label{fig: vis}
\end{figure}

We further evaluate AMC models under different metrics. 
In Table~\ref{tab:clickture_other_metric}, AMC Full model achieves obvious increases under all metrics. We show the ROC curves in Fig~\ref{fig: roc}. The AUC of AMC Full model has an increase of 3.4\% compared to the state-of-the-art method, which proves the effectiveness of the AMC learning method. Some visualization results are shown in Fig~\ref{fig: vis}.

\begin{table}[t]
  \centering
  \small
  \begin{tabular}{|l|c|c|c|c|c|} \hline
    Approach & R@1 & R@5 & R@10 & R@15 & R@20 \\ \hline
    DSSM-Img & 0.0767 & 0.2778 & 0.4025 & 0.4617 & 0.4891 \\ 
    DSSM-Key & 0.0980 & 0.3076 & 0.4207 & 0.4700 & 0.4926 \\ \hline
    Img\textsubscript{ATT} & 0.0782 & 0.2793 & 0.4049 & 0.4642 & 0.4918 \\
    Key\textsubscript{ATT} & 0.1042 & 0.3187 & 0.4322 & 0.4803 & 0.5019 \\ 
    Img\textsubscript{ATT}-Key\textsubscript{ATT}-LF & 0.1106 & 0.3445 & 0.4620 & 0.5108 & 0.5327 \\ 
    AMC Full & \textbf{0.1168} & \textbf{0.3504} & \textbf{0.4673} & \textbf{0.5148} & \textbf{0.5414} \\ \hline
  \end{tabular}
  \vspace{1.0mm}
\caption{Performance of different models on ASD. The evaluation metrics are R@1, 5, 10, 15, 20 (correspond to 2\textsuperscript{nd} to 6\textsuperscript{th} column).}\label{tab:rd_model}
\end{table}

\textbf{Performance on ASD.} We observe similar improvement by applying different attention mechanisms on AMC models in Table~\ref{tab:rd_model}. 
For intra-attention networks, LAN (Key\textsubscript{ATT}) achieves 0.6-1.2\% increase compared to DSSM-Key in R@k scores while VAN (Img\textsubscript{ATT}) does not observe much improvement ($\sim$0.2\% increase in R@k scores).
This is because most images in ASD contain only one object in the center, which takes 70\% of the space with clean backgrounds. 
In such case, VAN can offer limited boost in performance by focusing on informative regions.
We then combine VAN and LAN using LF. The resulting model, Img\textsubscript{ATT}-Key\textsubscript{ATT}-LF, achieves significant improvement in R@k scores, with 1.2-3.8\% increase compared to DSSM-Key and 3.2-6.5\% increase compared to DSSM-Img. 
We further apply the MTN to attend on different modalities, and get the AMC Full model. 
The AMC Full model achieves the best performance, with 0.6-1.0\% increase in R@k scores compared to late fusion model, 1.8-4.9\% increase in R@k scores compared to DSSM-Key and 3.8-7.1\% increase in R@k scores compared to DSSM-Img.

\textbf{Overfitting.} During training stage, we evaluate AMC models on test set every epoch. 
The training loss first reduces and converges at around epoch 12. 
The loss on test set follows the similar trend and converges at around epoch 14 on both Clickture and ASD, which indicates low possibility of overfitting.  
We further apply AMC models on caption ranking task which also achieves competitive performance.


\subsection{Caption ranking}\label{sec: caption rank}
\textbf{Experiment Setup.} For visual modality, we apply a pre-trained 200-layer Resnet~\cite{He2015} to extract image features as input. Each image is represented as a 2048D feature vector. 
To compare with \cite{lin2016leveraging}, we also extract image features using a pre-trained 19-layer VGG~\cite{Simonyan14c} network (4096D feature vector). 
For auto-tagged keywords, we remove stop words and uncommon words in the raw data, convert all words to lowercase, and tokenize each word to its index in the corresponding dictionary. 
The dictionary size for the keyword modality is 26,806. 
For caption modality, we extract skip-thought vectors~\cite{kiros2015skip} using a pre-trained model. Each caption is represented by a 4800D skip-thought vector. Same as \cite{kiros2015skip}, embedding vectors in all modalities are projected to 1000 dimensions ($d=1000$). The similarity between query and features from different modalities is measured by cosine distance in the AMC space.

\textbf{AMC models.} Same as the denotation in Sec~\ref{sec: multi-modal ir}, we apply latefusion (LF) and inter-attention (MTN) mechanisms to combine features from image modality and keyword modality (Key). Different AMC models' configuration is shown in Table~\ref{tab:amc_model_cap_retrieve}.

\begin{table}[t]
  \centering
  \begin{tabular}{|l|C{0.8cm}|C{0.8cm}|C{0.8cm}|C{0.8cm}|} \hline
    Approach & VGG & Res & LF & MTN \\ \hline
    Skip-Vgg~\cite{kiros2015skip} & \checkmark & & & \\ \hline
    Skip-Vgg-Key-LF & \checkmark & & \checkmark & \\ \hline
    AMC-Vgg & \checkmark & & & \checkmark \\ \hline
    Skip-Res & & \checkmark & & \\ \hline
    Skip-Res-Key-LF & & \checkmark & \checkmark & \\ \hline
    AMC-Res & & \checkmark & & \checkmark \\ \hline
  \end{tabular}
  \vspace{2.0mm}
\caption{Different models evaluated on CIC. Late fusion (LF) and inter-attention (MTN) networks are applied on multi-modalities. Caption modality is represented by Skip-thought vector (Skip). Image modality is represented by either VGG features (VGG) or Resnet features (Res).}\label{tab:amc_model_cap_retrieve}
\end{table}

\textbf{Training details.} We set margin $\alpha=0.2$ and number of negative samples $k=50$ for each correct caption-image (keyword) pair (Eq~\ref{equ: cap rank loss}).

\textbf{Evaluation Metric.} We follow the evaluation metric reported in \cite{karpathy2015deep}. Same as \cite{karpathy2015deep,kiros2014unifying,kiros2015skip,klein2015associating,ma2015multimodal,mao2014deep}, we report the caption retrieval performance on the first 1,000 test images. For a test image, the caption retrieval system needs to find any 1 out of its 5 candidate captions from all 5,000 test captions. We report recall@(1, 5, 10), which is the fraction of times a correct caption is found among the top (1, 5, 10) ranking results.

\textbf{Performance comparison.} AMC models provide very competitive results even without a complex language model, \emph{e.g.}, recurrent neural network (RNN), convolutional neural network (CNN) or Gaussian mixture models (GMM), to process captions compared to models in~\cite{karpathy2015deep,kiros2014unifying,kiros2015skip,klein2015associating,ma2015multimodal,mao2014deep}. 
In Table~\ref{tab:mic_results}, we first combine keyword and image modalities using latefusion (Skip-Vgg-Key-LF). Skip-Vgg-Key-LF gives small improvement in performance by $\mathtt{\sim}0.6\%$ in R@(1, 5, 10). 
This indicates that keyword modality provides useful information but further care is needed to put it to better use. Thus, we apply the inter-attention network (AMC-Vgg) to select informative modalities, which boosts the performance by a large margin, with 3.5\%, 1.9\% and 1.5\% increase in R@(1, 5, 10), respectively. 
We further change the image features to Resnet features, and observe similar performance improvement as Vgg features. 
The final model (AMC-Res), which applies MTN on Resnet-based image modality and keyword modality, achieves very close performance on R@1 as~\cite{ma2015multimodal}, on R@5 as~\cite{kiros2014unifying} and even surpasses the state-of-the-art result on R@10. 
We notice that AMC model does not achieve better results in R@5 compared to~\cite{mao2014deep,ma2015multimodal,kiros2014unifying}. 
This is because we adopt a relatively simple language model (Skip-thought vector~\cite{kiros2015skip}) for captions, with base performance at 33.5\% in R@5.   
Equipped with a more complex RNN / CNN model to process caption modality, AMC models will expect further boost in performance. 

\begin{table}[t]
  \centering
  \begin{tabular}{|l|c|c|c|} \hline
    Approach & R@1 & R@5 & R@10 \\ \hline
    Random & 0.1 & 0.5 & 1.0 \\
    DVSA~\cite{karpathy2015deep} & 38.4 & 69.9 & 80.5 \\
    FV~\cite{klein2015associating} & 39.4 & 67.9 & 80.5 \\
    $m$-RNN-vgg~\cite{mao2014deep} & 41.0 & 73.0 & 83.5 \\
    $m$-CNN\textsubscript{ENS}~\cite{ma2015multimodal} & 42.8 & 73.1 & 84.1 \\
    Kiros \emph{et al.}~\cite{kiros2014unifying} & \textbf{43.4} & \textbf{75.7} & 85.8 \\  \hline
    Skip-Vgg~\cite{kiros2015skip} & 33.5 & 68.6 & 81.5 \\ 
    Skip-Vgg-Key-LF & 34.2 & 69.3 & 82.0 \\
    AMC-Vgg & 37.0 & 70.5 & 83.0 \\
    Skip-Res & 39.5 & 73.6 & 86.1 \\ 
    Skip-Res-Key-LF & 40.1 & 74.2 & 86.5 \\
    AMC-Res & 41.4 & 75.1 & \textbf{87.8} \\ \hline
  \end{tabular}
  \vspace{1.5mm}
\caption{Performance of different models on CIC. The evaluation metrics are R@1, 5, 10(correspond to 2\textsuperscript{nd} to 4\textsuperscript{th} column). AMC models achieve competitive performance with only skip-thought vectors for caption modality among all VQA-agnostic models.}\label{tab:mic_results}
\end{table}

We notice that~\cite{lin2016leveraging} reports much better results on the caption ranking task compared to~\cite{karpathy2015deep,kiros2014unifying,kiros2015skip,klein2015associating,ma2015multimodal,mao2014deep}. However, the model in~\cite{lin2016leveraging} is called ``VQA-aware'' model, which encodes external VQA knowledge learned in the VQA task and fuses with the model in~\cite{kiros2014unifying}. AMC models, as well as models in~\cite{karpathy2015deep,kiros2014unifying,kiros2015skip,klein2015associating,ma2015multimodal,mao2014deep}, belong to ``VQA-agnostic'' models, which can be fused and enhanced by external VQA knowledge. We expect to see further boost in performance of AMC models on caption ranking task when the VQA knowledge data is made public.

\section{Conclusion}
We proposed an Attention guided Multi-modal Correlation (AMC) learning method. AMC models adaptively attend on useful modalities and filter out unrelated information within each modality according to the input query's intent. 
AMC framework can be further boosted by incorporating more image related modalities and external knowledge. This will be discussed in future work.

{\small
\bibliographystyle{ieee}
\bibliography{egbib}
}

\end{document}